\documentclass[10pt, a4paper]{article}
\usepackage{lrec}
\usepackage{graphicx}
\usepackage{tabularx}
\usepackage{soul}

\usepackage{epstopdf}
\usepackage[utf8]{inputenc}
\usepackage[russian,english]{babel}
\usepackage{hyperref}
\usepackage{xstring}
\usepackage{amsmath}

\title{MaSS: A Large and Clean Multilingual Corpus of Sentence-aligned Spoken Utterances Extracted from the Bible} 

\name{Marcely~Zanon~Boito$^{\ast1}$,  William N. Havard$^{\ast1,2}$, Mahault Garnerin$^{1,2}$,\\{\bf \large \'Eric Le~Ferrand$^{1}$, Laurent~Besacier$^{1}$}}
\address{
 $^1$LIG, Univ. Grenoble Alpes, CNRS, Grenoble INP, F-38000 Grenoble, France\\
 $^2$LIDILEM, Univ. Grenoble Alpes, F-38000 Grenoble, France\\
 \{first.last-name\}@univ-grenoble-alpes.fr\\
 $^{\ast}$Both authors have contributed equally to this paper.
}

\abstract{
The \textit{CMU Wilderness Multilingual Speech Dataset}~\cite{Black_CMU}
is a newly published multilingual speech dataset based on
recorded readings of the New Testament. It provides data to build Automatic Speech Recognition (ASR) and Text-to-Speech (TTS) models for potentially 700 languages. However, the fact that the source content (the Bible) is the same for all the languages is not exploited to date. Therefore, this article proposes to add multilingual links between speech segments in different languages, and shares a large and clean dataset of 8,130 parallel spoken utterances across 8 languages (56 language pairs). We name this corpus MaSS (\textbf{M}ultilingu\textbf{a}l corpus of \textbf{S}entence-aligned \textbf{S}poken utterances). The covered languages (Basque, English, Finnish, French, Hungarian, Romanian, Russian and Spanish) allow researches on speech-to-speech alignment as well as on translation for typologically different language pairs. The quality of the final corpus is attested by human evaluation performed on a corpus subset (100 utterances, 8 language pairs). Lastly, we showcase the usefulness of the final product on a bilingual speech retrieval task.
\\ \newline  }

\begin{document}

\maketitleabstract
\section{Introduction}
Recently, a remarkable work introduced the \textit{CMU Wilderness Multilingual Speech Dataset}~\cite{Black_CMU}.\footnote{Available at \url{http://www.festvox.org/cmu_wilderness/index.html}} Based on readings of the New Testament from \textit{The Faith Comes By Hearing} website, it provides data to build Automatic-Speech-Recognition (ASR) and Text-to-Speech~(TTS) models for potentially 700 languages. Such a resource allows the community to experiment and to develop speech technologies on an unprecedented number of languages. However, the fact that the initial language material from these monolingual corpora (the Bible) is the same for all languages, thus constituting a multilingual and comparable\footnote{Our definition of a \textit{comparable} corpus in this work is the following: a non-sentence-aligned corpus, parallel at a broader granularity (e.g. chapter, document).} spoken corpus, is not exploited to date.

Therefore, this article proposes an automatic pipeline for adding multilingual links between small speech segments in different languages. We apply our method to 8 languages (Basque, English, Finnish, French, Hungarian, Romanian, Russian and Spanish), resulting in 56 language pairs for which we obtain speech-to-speech, speech-to-text and text-to-text alignments.
In order to ensure the quality of the pipeline, a human evaluation was performed on a corpus subset (8 language pairs, 100 sentences) by bilingual native speakers. The current version of our dataset (named MaSS for \textbf{M}ultilingu\textbf{a}l corpus of \textbf{S}entence-aligned \textbf{S}poken utterances) is 
freely available to the community, together with instructions and scripts allowing the pipeline extension to new languages.\footnote{Available at \label{link_dataset}\url{https://github.com/getalp/mass-dataset}}

We believe the obtained corpus can be useful in several applications, such as speech-to-speech retrieval~\cite{lee2015spoken}, multilingual speech representation learning~\cite{Harwath18_interlingua} and direct speech-to-speech translation (so far, mostly direct speech-to-text translation has been investigated~\cite{berard-nips2016,weiss2017sequence,bansal2017,berard:hal-01709586}). Moreover, typological and dialectal fields 
could use such a corpus to solve some of the following novel tasks using parallel speech: word alignment, bilingual lexicon extraction, and semantic retrieval.

This paper is organized as follows: after briefly presenting related works in Section 
2, we review the dataset source and extraction pipeline in Section 
3. Section 
4 describes the human verification performed and comments on some of the linguistic features present in the covered languages. Section 
5 presents a possible application of the dataset: speech-to-speech retrieval. Section 
6 
concludes this work.

\section{Related Work}
\label{sec:related}

\subsection{End-to-end Speech Translation}
Previous Automatic Speech-to-Text Translation~(AST) systems operate in two steps:  source language Automatic Speech Recognition~(ASR) and source-to-target text Machine Translation~(MT). However, recent works have attempted to build end-to-end AST without using source language transcription during learning or decoding~\cite{berard-nips2016,weiss2017sequence}, or by using it at training time only~\cite{berard:hal-01709586}. Very recently several extensions of these pioneering works were introduced: low-resource AST~\cite{DBLP:journals/corr/abs-1809-01431}, unsupervised AST~\cite{DBLP:journals/corr/abs-1811-01307}, end-to-end speech-to-speech translation~(\textit{Translatotron})~\cite{DBLP:journals/corr/abs-1904-06037}. Improvements for 
end-to-end AST were also proposed by using weakly supervised data~\cite{DBLP:journals/corr/abs-1811-02050}, or by adding a second attention mechanism~\cite{DBLP:journals/corr/abs-1904-07209}.

\subsection{Multilingual Approaches}
Multilingual approaches for speech and language processing are growing 
ever more popular. They are made possible by the availability of massively parallel language resources covering an increasing number of languages of the world. These resources feed truly multilingual approaches, such as machine translation~\cite{DBLP:journals/corr/abs-1903-00089}, syntax parsing~\cite{Nivre2016}, automatic speech recognition~\cite{Schultz2014,Adams18_naacl}, lexical disambiguation~\cite{Navigli2010,serasset2015dbnary}, and computational dialectology~\cite{Christodoulopoulos2015}. 

\subsection{Corpora for End-to-end Speech Translation}
To date, few  datasets are  available for  multilingual automatic speech translation (only
a few parallel corpora publicly available\footnote{Table 1 in~\cite{mustc19} provides a good survey.}). For instance, \textit{Fisher} and \textit{Callhome} Spanish-English corpora~\cite{fishercorpus} provide 38 hours of speech transcriptions of telephonic conversations aligned with their translations. However, these corpora are only medium size and contain low-bandwidth recordings. Microsoft Speech Language Translation (MSLT) corpus~\cite{Federmann2016MicrosoftSL} also provides speech aligned to translated text, but this corpus is rather small (less than 8 hours per language). A 236 hours extension of \textit{Librispeech} with French translations was proposed by~\newcite{DBLP:journals/corr/abs-1802-03142}. They exploited automatic alignment procedures, first at the text level (between transcriptions and translations), and then between the text and the corresponding audio segments.

Inspired by this work, \newcite{mustc19} created ~MuST-C, a multilingual speech translation corpus for training end-to-end AST systems from English into 8 languages.\footnote{Available at \url{https://ict.fbk.eu/must-c}} Similar in size, the English-Portuguese dataset \textit{How2}~\cite{sanabria18how2} was created by translating English short tutorials into Portuguese using a crowd-sourcing platform. More recently, \newcite{iranzo2020europarlst} introduced a multilingual speech corpus including several source languages.
The remark that can be made on all these corpora is that they are limited to Indo-European languages and thus typologically similar.


\section{A Large and Clean Subset of Sentence Aligned Spoken Utterances (MaSS)}
\label{sec:corpus}
In this section we present the source material for our multilingual corpus (Section~\ref{sec:bible}), we briefly explain the CMU speech-to-text pipeline (Section~\ref{sec:cmu}), and we detail our speech-to-speech pipeline (Section~\ref{sec:multilingual}).

\subsection{The Source Material: Bible.is}\label{sec:bible}

The \textit{Faith Comes By Hearing} website\footnote{Available at \url{https://www.bible.is}} (or simply \textit{bible.is}) is an online platform that provides audio-books of the Bible with transcriptions in 1,294 languages. These recordings are a collection of field, virtual and partner recordings. 
In all cases, only native speakers participate in the recordings, and the number of different voices can go from one up to twenty five. 
Moreover, the recordings can be performed in \textit{drama} and \textit{non-drama} fashion, the former being an acted version of the text, 
corresponding to less tailored realizations. Finally, based on exchanges with the target users (the native community), background music can be added to the recordings.\footnote{More information available at \url{https://www.faithcomesbyhearing.com/mission/recordings}} In summary, while the written content is always the same across different languages, the corresponding speech can be quite different in terms of realization (drama and non-drama), number of speakers, acoustic quality (field, virtual or partner recordings), and can sometimes contain background noise (music).

\subsection{The CMU Wilderness Multilingual Speech Corpus}\label{sec:cmu}
The CMU Wilderness corpus~\cite{Black_CMU} is a speech dataset containing over 700 different languages for which it provides audio excerpts aligned with their transcription. Each language accounts for around 20 hours of data 
extracted from readings of the New Testament, and available at the \textit{bible.is} website. 
Segmentation was made at the sentence level, and alignment between speech and corresponding text can be 
obtained with the pipeline provided along with the dataset. This pipeline, notably, 
can process a large amount of languages without using any extra resources such as acoustic models or pronunciation dictionaries. 

However, for most of the languages on the website, several recording versions are available, each of them having significant differences in speech content, as explained in Section 
3.1. As this pipeline extracted the soundtracks from the defaults links, audio excerpts often contain music, and it is unknown if drama or non-drama versions were selected. Thus, although the quality of the alignment is good for many languages, it could be inaccurate (or noisy) for an unknown subset.

Lastly, the final segmentation from chapters was obtained through the use of punctuation marks. While efficient for a speech-to-text monolingual scenario, this strategy does not allow accurate multilingual alignment, since different languages and translations may result in different sentence segmentation and ordering.

\subsection{Our Pipeline: from Speech-to-text to Speech-to-speech Alignment}\label{sec:multilingual}
\begin{figure}
\centering
\includegraphics[scale=0.53]{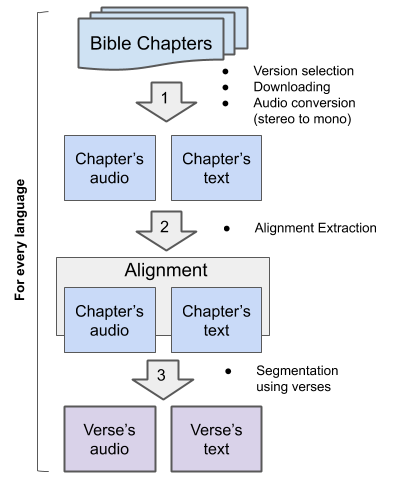}
    \caption{The pipeline for a given language in the \textit{bible.is} website.}
\label{pipelineexample}
\end{figure}

As far as multilingual alignment is concerned, Bible chapters are inherently aligned at the \textit{chapter} level. But Bible chapters are 
very long excerpts, with an average duration of 5 minutes. Alignments at this broad granularity are not relevant for research in speech-to-speech translation or speech-to-speech alignment.
Thus, we propose a new extraction methodology that allows us to obtain fully aligned speech segments at a much smaller granularity (segments between 8 to 10 seconds).
Our pipeline is summarized in Figure~\ref{pipelineexample} and described below.

\subsubsection{Alignment pipeline}
\textbf{1. Extracting clean spoken chapters.} Starting from the pipeline described in the last section, which provides scripts for downloading audio data and transcriptions from the \textit{bible.is} website, we downloaded all the 260 chapters from the New Testament in several languages. We selected (after having manually sampled the website) \textit{non-drama} 
versions (as opposed to \textit{drama}) that contain standard speech and pronunciation, and mostly, no background music. The audios are also converted from stereo to mono for the purpose of the following steps.

\textbf{2. Aligning speech and text for each chapter.} For each chapter, we extracted speech-to-text alignments through the \textit{Maus} forced aligner\footnote{Available at \url{https://clarin.phonetik.uni-muenchen.de/BASWebServices/interface/WebMAUSBasic}}~\cite{kisler2017multilingual} online platform. During this step, we kept languages with good audio quality and for which an acoustic model was available in the off-the-shelf forced aligner tool. Our final set was reduced to the following eight languages: Basque, English, Finnish, French, Hungarian, Romanian, Russian and Spanish.

\textbf{3. Segmenting chapters into verses.}
 Any written chapter of the Bible is inherently segmented into \textit{verses}. A \textit{verse} is the minimal segmentation unit used in the Bible and corresponds to a sentence, or more rarely to a phrase or a clause. In order to segment our audio files in such smaller units, we aligned our \textit{TextGrid files} (from step 2) with a written version of the Bible containing \textit{verse} information. This alignment is rather trivial, since, after removing punctuation, both texts have the same content. 
 After this step, all audio chapters are segmented into verses and receive \textit{IDs} based on their English chapter name, 
and their verse number (e.g. ``Matthew\_chapter1\_verse3'').

\subsubsection{Result and Comparison}

Considering that all chapters consist of the same set of verses, the verse numbers give us a multilingual alignment between all verses for all the language pairs.\footnote{This is mostly true, but for a small subset of chapters, due to different Bible versions and different translation approaches, the number of aligned speech verses will differ slightly.}
Thus, the output of our pipeline is a set of 8,160 audios segments, aligned at verse-level, in eight different languages, with an average of 20 hours of speech for each language.
Finally, corpus statistics are presented in Table~\ref{tablestats}. 
\begin{table*}[hbt!]
    
    \resizebox{\textwidth}{!}{
        \begin{tabular}{l|rrccc|cc}
        \hline
        \textbf{Languages} & \textbf{\# types} & \textbf{\# tokens} & \textbf{Types per verse} & \textbf{Tokens per verse} & \textbf{Avg. token length} & \textbf{Audio length (h)} & \textbf{Avg. verse length (s)} \\ \hline
        (EN) English    & 6,471   & 176,461  & 18.03           & 21.52            & 3.82              & 18.50            & 8.27            \\
        (ES) Spanish    & 11,903  & 168,255  & 17.90            & 20.52            & 4.17              & 21.49           & 9.58            \\
        (EU) Basque     & 14,514  & 128,946  & 14.88           & 15.78            & 5.55              & 22.76           & 9.75           \\
        (FI) Finnish   & 18,824  & 134,827  & 15.04           & 16.44            & 5.66              & 23.16           & 10.21           \\
        (FR) French    & 10,080  & 183,786  & 19.25           & 22.36            & 4.02              & 19.41           & 8.62            \\
        (HU) Hungarian  & 20,457  & 135,254  & 15.01           & 16.46            & 5.07              & 21.12           & 9.29            \\
        (RO) Romanian  & 9,581   & 169,328  & 18.19           & 20.61            & 4.14              & 23.11           & 10.16           \\
        (RU) Russian    & 16,758  & 129,973  & 14.50            & 15.82            & 4.44              & 22.90            & 9.70             \\
        \hline
        \end{tabular}
    }
    \caption{Statistics of the MaSS corpus.}
    \label{tablestats}
\end{table*}

For justifying the need of extending the approach presented in Section 
3.2, 
Table~\ref{alignmentcomparison} presents a comparison between our corpus (bottom) output and theirs (top). This comparison takes the speech file numbering on their pipeline as multilingual alignment clue, since no other information is 
available. We can observe that by segmenting based on punctuation, the multilingual alignment quickly becomes incorrect: the segmentation on the third file, based on a punctuation mark not present in the English text, shifts the alignment for the rest of the chapter.

\begin{table*}[hbt!]

\large
\resizebox{\textwidth}{!}{
\begin{tabular}{m{1.3cm}|m{10cm}|m{10cm}}
\hline
\multicolumn{3}{c}{\textbf{ Alignment from \newcite{Black_CMU}}}                                                                                                                                                                                                                                                                                                                                                                                                       \\ 
\textbf{Files}        & \textbf{French}                                                                                                                                                                                                                     & \textbf{English}                                                                                                                                                                                               \\ \hline
00001 & Matthieu                                                                                                                                                                                                                 & Matthew                                                                                                                                                                                             \\ \hline
00002 & J\'esus descend de la montagne et des foules nombreuses le suivent.                                                                                                            & When he came down from the mountainside, large crowds followed him.                                                                                   \\ \hline
00003 & Un l\'epreux s'approche, il se met \`a genoux  devant J\'esus et lui dit :                                                                                                  & A man with leprosy came and knelt before him and said, \textit{``Lord, if you are willing, you can make me clean."}                                            \\ \hline
00004 & Seigneur, si tu le veux, tu peux me gu\'erir !                                                                                                                                                                              & \textit{Jesus reached out his hand and touched the man. ``I am willing," he said. ``Be clean!" Immediately he was cured of his leprosy.}                        \\ \hline 
                                                                                              
\multicolumn{3}{c}{\textbf{Our alignment}}                                                                                                                                                                                                                                                                                                                                                                                                          \\ 
\textbf{Verses}       & \textbf{French}                                                                                                                                                                                                                     & \textbf{English}                                                                                                                                                                                               \\ \hline
00    & Matthieu 8                                                                                                                                                                                                                 & Matthew 8                                                                                                                                                                                             \\ \hline
01    & Lorsque J\'esus fut descendu de la montagne une grande foule le suivit                                                                                                        & When he came down from the mountain  great crowds followed him                                                                                          \\ \hline
02    & Et voici un l\'epreux s'\'etant approch\'e se prosterna devant lui et dit : Seigneur si tu le veux tu peux me rendre pur                                                           & And behold a leper came to him and knelt before him saying Lord if you will you can make me clean                                                         \\ \hline
03    & J\'esus \'etendit la main le toucha et dit : Je le veux sois pur Aussit\^ot il fut purifi\'e de sa l\`epre                                                                               & And Jesus stretched out his hand and touched him saying I will be clean And immediately his leprosy was cleansed                             
\\ \hline
\end{tabular}
}
\caption{A comparison between CMU's multilingual alignment and ours. Text in italic shows alignment mismatches between English and French. We used a slightly different (\textit{non-drama}) version of the Bible, hence the small differences in the displayed texts.}
\label{alignmentcomparison}
\end{table*}

\subsubsection{Reproducibility}

The presented pipeline performs automatic verse-level alignment using Bible chapters. All the scripts used in this work are available, together with the resulting dataset.\textsuperscript{\ref{link_dataset}} For extending it to a new language, here are some recommendations:

\begin{itemize}
    \item \textbf{Bible version:} as discussed in Section 
    3.1, a language can have several versions available on the website. For ensuring the best quality possible, manual inspection in one chapter can be quickly performed to identify a non-drama 
    version, but it is not mandatory.
    \item \textbf{Alignment Tool:} for generating verse-level alignment, 
    a chapter-level alignment between speech and text is needed. While we use the \textit{Maus forced aligner} for this task, any aligner able to provide a \textit{TextGrid file} as output can be used at this stage.
\end{itemize}

\section{Resource Evaluation and Analysis}
\label{sec:validation}

\subsection{Human Evaluation: Speech Alignment Quality}
Having obtained multilingual alignments between spoken utterances, we attest their quality by performing a human evaluation on a corpus subset, covering the eight language pairs for which we were able to find bilingual judges.

We implemented an online evaluation platform with 100 randomly selected verses in these 8 different language pairs. Judges were asked to evaluate the spoken alignments using a scale from 1 to 5 (1 meaning the two audio excerpts do not have any information in common, and 5 meaning they are perfectly aligned). Aiming at the most uniform evaluation possible, we provided guidelines and examples to our evaluators. Transcriptions were also displayed as a cognitive support in evaluation.

The eight language pairs are the following: French-English~(FR-EN), French-Spanish~(FR-ES), French-Romanian~(FR-RO), English-Spanish~(EN-ES), English-Finnish~(EN-FI), English-Hungarian~(EN-HU), English-Romanian~(EN-RO) and English-Russian (EN-RU). This selection is a trade-off between the difficulty of finding judges and the desire to provide a good typological variety in our evaluation data. Basque was also chosen due to the fact it is language isolate, that is, a language that has no known connection to any other language. However, we were unable to find judges to perform the evaluation on any language pair including it.

Table~\ref{tab:eval_summ} summarizes the results of the human evaluation. Evaluation scores are good, with a mean value of 4.41. Moreover, for every language pair evaluated (except for FR-ES), the median score is the maximum score, hence confirming the quality of the alignment. However, when trying to quantify rater's agreement, we obtained mixed results. Percentage of agreement with tolerance 1 (meaning raters differing by one-scale degree are interpreted as agreeing) varies from 59.6\% (EN-RO) to 95.96\% (EN-HU).

\begin{table}[hbt!]
\centering

\begin{tabular}{lcccccc}
    \hline
        & \=x    &  $\sigma$    & \textbf{med}   & \textbf{min}   & \textbf{max} & \textbf{\# Eval.} \\ 
        \hline
    EN - ES & 4.56 & 0.62 & 5     & 3     & 5   & 2  \\
    EN - FI  & 4.37 & 0.92 & 5     & 1     & 5   & 1  \\
    EN - HU  & 4.44 & 0.88 & 5     & 1     & 5   & 2  \\
    EN - RO  & 4.24 & 0.97 & 5     & 1     & 5   & 6  \\
    EN - RU  & 4.56 & 0.83 & 5     & 1     & 5   & 3  \\
    FR - EN  & 4.38 & 0.79 & 5     & 1     & 5   & 5  \\    
    FR - ES  & 4.22 & 0.89 & 4     & 2     & 5   & 2  \\
    FR - RO  & 4.51 & 0.90 & 5     & 1     & 5   & 1  \\   
    \hline
    \textbf{All} & \textbf{4.36} & \textbf{0.88} & \textbf{5} & \textbf{1} & \textbf{5}   & \textbf{22} \\
\hline
\end{tabular}
\caption{Result of the manual inspection of the speech alignment quality performed on 8 language pairs (100 sentences). Scale is from 1 to 5 (higher is better). Last column refers to the number of evaluators for a given language pair.}
\label{tab:eval_summ}
\end{table}

\subsection{Corpus Linguistic Analysis}
\label{sec:overview}

Regarding content, the corpus features languages belonging to different families. These are listed as follows:
\begin{itemize}
  \item Indo-European:
    \begin{itemize}
      \item Romance: \textbf{French}, \textbf{Romanian}, \textbf{Spanish}
      \vspace{-0.15cm}
      \item Germanic: \textbf{English}
      \vspace{-0.15cm}
      \item Slavic: \textbf{Russian}
     \end{itemize}
  \item Uralic:
    \begin{itemize}
      \item Ugric: \textbf{Hungarian}
      \vspace{-0.15cm}
      \item Finnic: \textbf{Finnish}
     \end{itemize}
  \item Language Isolate: \textbf{Basque}
\end{itemize}

It should be noted that these languages are very different from a typological point of view. First of all, Basque, Finnish, Hungarian, Romanian and Russian mainly use case marking to indicate the function of a word\footnote{Case markers are small grammatical morphemes added to a word to indicate its grammatical function (eg. subject, object, etc.) within a clause/sentence.} in a sentence, while English, French and Spanish rely on word position and prepositions for the same purposes. Basque, Finnish and Hungarian are agglutinative languages, while English, French, Romanian, Russian and Spanish are fusional languages. Thus, for the former group, grammatical markers will bear only one meaning, while in the latter, grammatical markers will bear several meanings at the same time.\footnote{Compare Hungarian ``h\'az-ak-nak" (house-\textsc{Pl-Dat}) and Russian ``\foreignlanguage{russian}{дом-ам}" (house-\textsc{Pl.Dat}). Words in agglutinative languages are comparatively longer than their equivalent in fusional languages.}

Basque is even more special as this language features ergative-absolutive marking while the other languages use nominative-accusative marking. 
In languages using ergative-absolutive marking, the subject of an intransitive verb and the patient of a transitive verb are treated alike and receive the same case marker, while the agent of a transitive verb is treated differently than the subject of an intransitive verb. Romanian also presents an interesting morphological characteristic regarding determiners: the definite article is suffixed to the word whereas indefinite articles are usually prefixed, for instance: ``un-băiat" (INDEF-boy: ``a boy") and ``băiat-ul" (boy-DEF: ``the boy"). Finnish and Russian on the other hand do not have any article, neither definite nor indefinite.

Another interesting linguistic phenomenon to observe is the existence of grammatical genders. Russian features three genders (feminine, masculine and neutral) whereas French features only two (feminine and masculine), and Basque and Finnish present no grammatical genders at all. From a syntactic point of view, English, French and Spanish have a relatively fixed word order (and mainly follow the Subject-Verb-Object~(SVO) pattern), while word order is more flexible in Basque, Finnish, Hungarian, Romanian and Russian, mainly due to the fact that these languages use case markers.

Due to all the diverse linguistic features described in this section, we believe this dataset could be used for a wide variety of tasks, such as natural language grammar induction from raw speech, automatic typological features retrieval, speech-to-speech translation, and speech-to-speech retrieval. The latter is illustrated on  Section 
5. Moreover, this dataset could also serve as a benchmark for evaluating computational language documentation techniques that work on speech inputs. 

\section{Use Case: Multilingual Speech Retrieval Task Baseline}
\label{sec:example}

In this section we showcase the usefulness of our corpus on a multilingual setting. We perform speech-to-speech retrieval by adapting a model for visually grounded speech~\cite{conf/eccv/HarwathRSCTG18}, and we discuss the results for our \textit{baseline model}.

\subsection{Task and Model Definition}
For performing multilingual speech retrieval, we adapted the model\footnote{Available at \url{https://github.com/dharwath/DAVEnet-pytorch}} proposed by~\newcite{conf/eccv/HarwathRSCTG18}. This model was primarily designed to retrieve images from speech utterances, and it is made of two networks: a speech and a image encoder. By projecting both representations to the same shared space, the model is thus able to learn the relationship between speech segments and the image contents. For our speech-to-speech task, we replaced the image encoder by a (second) speech encoder.\footnote{Modified code available at \url{https://github.com/getalp/BibleNet}} 

Both speech encoders consist of a convolution bank~\cite{tacotron_v1} followed by two layers of bidirectional LSTM~\cite{LSTM}, and of an attention mechanism~\cite{bahdanau2014neural} which computes a weighted average of the LSTM's activations. The convolution bank consists of a set of $K=16$ 1D-convolution filters, where the $k^{th}$ convolution has a kernel of width $k$. Each convolution filter consists of 40 units with ReLU activation and stride of 1. The batch-normed output of each convolution is then stacked and the resulting matrix is linearly projected to fit the LSTM's input dimension of size 256. 

Our model's inputs are 
Mel filterbank spectrograms (40 mel coefficients with a Hamming window size of 25ms and stride of 10ms) extracted from raw speech. The network is trained to minimize the contrastive loss function in Equation~\ref{eq:loss_func}, which minimizes the cosine distance $d$ between a verse in a given language $A$, and its corresponding verse in a given language $B$. It does so by maximizing the distance between mismatching verses pairs (with a given margin $\alpha$). Thus, verses corresponding to direct translations should lie close in the embedding space. Finally, contrary to~\newcite{Harwath18_interlingua}, in which only one negative example for caption is sampled, 
we adopted the 
method from~\newcite{Chrupala2017}, 
considering every other verse in the batch as a negative example.

\begin{equation}
  \resizebox{1\hsize}{!}{
  \begin{math}
    \begin{split}
    L(v_A, v_B, \alpha) =
      \sum_{v_A, v_B} 
      &\Bigg( \sum_{v_A'}\max [0, \alpha + d(v_A,v_B) - d(v_A',v_B)]\\
      &+\sum_{v_B'}\max[0, \alpha + d(v_A,v_B) - d(v_A,v_B')] \Bigg)
    \end{split}
  \end{math}
  }
  \label{eq:loss_func}
\end{equation}

\subsection{Results}       
We trained an instance of this model for seven language pairs, always keeping English as source language. The 8,160 common verses were randomly split between train (80\%), validation (10\%) and test (10\%) sets. Batches were of size 16, and models were all trained for 100 epochs. Table~\ref{table:speech-2-speech} presents our results for the retrieval task.

\begin{table}[!htbp]
    \centering
     
    \begin{tabular}{lcccc}
        \hline
        \textbf{Query}   & \textbf{R@1} & \textbf{R@5} & \textbf{R@10} & \textbf{$\widetilde{r}$} \\
        \hline
        \textbf{EN-EU} & 0.173 & 0.395 & 0.523 & 9   \\
        \textbf{EN-ES} & 0.130 & 0.341 & 0.469 & 12   \\
        \textbf{EN-HU} & 0.116 & 0.319 & 0.455 & 13   \\
        \textbf{EN-RU} & 0.102 & 0.308 & 0.414 & 16   \\
        \textbf{EN-RO} & 0.085 & 0.289 & 0.396 & 17   \\
        \textbf{EN-FR} & 0.092 & 0.259 & 0.364 & 22   \\
        \textbf{EN-FI} & 0.076 & 0.202 & 0.293 & 26   \\
        \hline
    \end{tabular}
    \caption{Recall at top 1, 5, and 10 retrieval. Median rank \textbf{$\widetilde{r}$} on a verse-to-verse retrieval task is also provided. Results are reported on the test set (816 verses).
     Chance recalls are 0.001 (R@1), 0.006 (R@5) and 0.012 (R@10). Chance median $\widetilde{r}$ is 408.5. }
     \label{table:speech-2-speech}
\end{table}

Results show that, while such a speech-to-speech task is challenging, it is possible to obtain bilingual speech embeddings that perform reasonably well on a multilingual retrieval task. The recall and rank results are far above the chance values. We also scored a simple baseline that uses utterance length to retrieve spoken verses (in other words, it uses only distance between spoken utterances' lengths  to solve the retrieval task). With this baseline, medium ranks are better than chance level ($\widetilde{r}=408$) but vary from $\widetilde{r}=136$~(EN-FR) to $\widetilde{r}=219$~(EN-FI), which is very poor compared to our baseline model.
Interestingly, our best results, obtained for EN-EU ($\widetilde{r}=9$) and EN-ES~($\widetilde{r}=12$), illustrate that speech-to-speech retrieval task is feasible even for pairs of typologically different languages.

Following this experiment, we investigated the correlation between the median rank and two variables: the quality of the alignment (human evaluation) and the syntactic distance between the language pairs (using the \textit{lang2vec} library \cite{littell2017uriel}). Results are provided at Table~\ref{eval_lang2vec}.
While there is no correlation between the rank and the syntactic distance, there is a strong negative correlation with respect to the human evaluation (significant for $p<0.1$). 
One possible explanation for this result may be that higher quality alignments (measured by the human evaluation~$\widetilde{x}$) lead to a slightly easier corpus for the speech-speech retrieval task (difficulty being measured by the rank $\widetilde{r}$). If confirmed, this result would suggest that speech-to-speech retrieval scores 
are a good proxy for rating alignment corpus quality, as performed for text by~\newcite{DBLP:journals/corr/abs-1907-05791} through the use of NMT. 
\begin{table}[hbt!]
\centering

\begin{tabular}{l|cccc}
\hline
\textbf{Languages} & \textbf{$\widetilde{r}$}         & \textbf{Quality (\=x)}  & \textbf{Syntactic dist.} \\\hline
EN - EU          & 9           & \textit{NA}        & 0.61   \\
EN - ES          & 12            & 4.56      & 0.40   \\
EN - HU          & 13           & 4.44       & 0.57   \\
EN - RU          & 16           & 4.56        & 0.49   \\
EN - RO          & 17           & 4.51       & 0.53   \\
EN - FR          & 22           & 4.38      & 0.46   \\
EN - FI          & 26           & 4.37       & 0.53   \\\hline

\multicolumn{2}{l}{Correlation} & -0.76  & -0.21 \\
\hline
\end{tabular}
\caption{Correlation between median rank and 1) alignment quality (from manual evaluation) 2) syntactic distance between languages (measured with \textit{lang2vec}).}
\label{eval_lang2vec}
\end{table}

\section{Conclusion}
\label{sec:conclusion}
In this paper, we presented the creation of an automatically generated clean and controlled parallel corpus based on the \textit{CMU Wilderness Multilingual Speech Dataset}. Our resource, called MaSS, contains 20 hours of speech in eight languages (Basque, English, Finnish, French, Hungarian, Romanian, Russian and Spanish) and presents both speech-to-text and speech-to-speech alignments. The quality of the corpus was verified on a subset of 100 sentences in 8 language pairs by native speakers. The pipeline used for creating this dataset, as well as the computed forced alignments for each of the chosen languages, are openly accessible.\textsuperscript{\ref{link_dataset}} Only eight languages are currently covered, but we believe the same methodology could easily be applied for extending it to new languages.

\section{Bibliographical References}
\label{main:ref}

\bibliographystyle{lrec}
\bibliography{merge_bib}


\end{document}